\documentclass{article}
\usepackage[utf8]{inputenc}
\usepackage[a4paper, margin={1in,1in}]{geometry}
\usepackage{hyperref}
\usepackage{graphicx}
\usepackage{caption}
\usepackage{amsmath}
\usepackage{amssymb}
\hypersetup{
    colorlinks=true,
    linkcolor=blue,
    filecolor=magenta,      
    urlcolor=blue,
    pdfpagemode=FullScreen,
    }

\title{\textbf{A visual introduction to Gaussian Belief Propagation}
\\ \Large A Framework for Distributed Inference with Emerging Hardware. 
\\\vspace{0.2in}
\url{https://gaussianbp.github.io}
\\\vspace{0.2in}
}
\author{
    Joseph Ortiz$^{1}$, Talfan Evans$^{1, 2}$, Andrew J. Davison$^{1}$ \\
    $^{1}$Imperial College London, $^{2}$DeepMind \\
    {\normalsize \texttt{joeaortiz16@gmail.com}} \\ \\
    
}
\date{June 21, 2021}

\begin{document}

\maketitle

{\large \textbf{\textcolor{red}{This is a citable PDF version of the online interactive article hosted at: \newline\url{https://gaussianbp.github.io}. This version is purely for citation purposes and the article should be read in the online format. In the online version, all figures are interactive figures and in this PDF version we include screenshots of the interactive figures.}}}

\begin{abstract}
    In this article, we present a visual introduction to Gaussian Belief Propagation (GBP), an approximate probabilistic inference algorithm that operates by passing messages between the nodes of arbitrarily structured factor graphs. A special case of loopy belief propagation, GBP updates rely only on local information and will converge independently of the message schedule. Our key argument is that, given recent trends in computing hardware, GBP has the right computational properties to act as a scalable distributed probabilistic inference framework for future machine learning systems.
\end{abstract}

\begin{figure}[h!]
    \centering
    \includegraphics[width=0.5\linewidth]{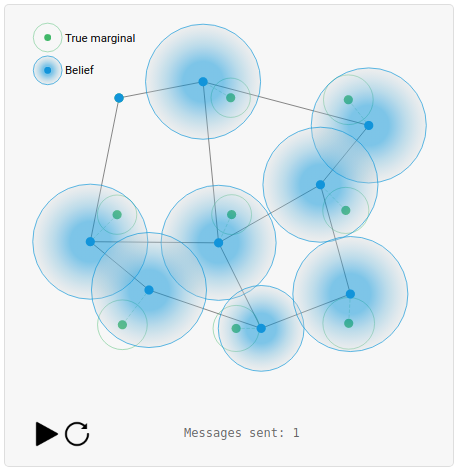}
    \caption{
        \textcolor{red}{INTERACTIVE FIGURE -- view at \url{https://gaussianbp.github.io}.}
        Gaussian Belief Propagation performs probabilistic inference iteratively and is convergent even when messages are passed randomly through the graph.
        Here GBP is applied to a geometric grid alignment problem.
        See an \hyperref[fig:gbpintuition]{interactive version} of this figure later in the article. 
    }
    \label{fig:teaser}
\end{figure}

\section{Introduction}

\subsection{Bayesian Probabilistic Inference}

Bayesian probability theory is the fundamental framework for dealing with uncertain data, and is at the core of practical systems in machine learning and robotics \cite{Ghahramani:Nature2015, davison2018futuremapping}. 
A probabilistic model relates unknown variables of interest to observable, known or assumed quantities and most generally takes the form of a graph whose connections encode those relationships.
Inference is the process of forming the posterior distribution to determine properties of the unknown variables, given the observations, such as their most probable values or their full marginal distributions.

There are various possible algorithms for probabilistic inference, many of which take advantage of specific problem structure for fast performance.
Efficient inference on models represented by large, dynamic and highly inter-connected graphs however remains computationally challenging and is already a limiting factor in real embodied systems.
Inference on such graphs will only become more important as general purpose perception systems strive towards more heterogeneous and dynamic representations containing abstractions (such as objects or concepts) with continually changing relationships \cite{bengio2017consciousness, davison2018futuremapping}
\footnote{
In AI, there has been a recent trend towards sparse representations and graphs which are well-suited for representing sparse high dimensional data.
Part of this trend has been driven by the fact that graphs are a natural representation in certain domains, and evidence of this is the rise of graph neural networks and graph-based probabilistic inference frameworks \cite{dellaert2017factor, davison2018futuremapping}.
Another trend is driven by the idea that massive sparsity is required to scale AI compute and evidence of this is the scaling of neural networks with gated computation \cite{shazeer2017outrageously, fedus2021switch} and the development of novel AI processors specifically for high performance on sparse representations \cite{Lacey:IPUBenchmarks2019}.
We believe that inference on large scale and dynamic probabilistic graphs will be an important part of this trend towards sparse computing to scale AI compute.}.

\subsection{The Bitter Lesson}

As we search for the best algorithms for probabilistic inference, we should recall the ``Bitter Lesson" of machine learning (ML) research: \textit{``general methods that leverage computation are ultimately the most effective, and by a large margin"} \cite{Sutton:BitterLesson2019}. 
A reminder of this is the success of CNN-driven deep learning, which is well suited to GPUs, the most powerful widely-available processors in recent years.

Looking to the future, we anticipate a long-term trend towards a ``hardware jungle" \cite{Sutter:Jungle2011} of parallel, heterogeneous, distributed and asynchronous computing systems which communicate in a peer-to-peer manner.
This will be at several levels: across networks of multiple smart devices operating in the same environment; across the many sensors, actuators and processors within individual embodied devices; and even within single processor chips themselves
\footnote{For single processor chips, a trend towards multicore designs with distributed on-core memory and ad-hoc communication is rapidly emerging \cite{Lacey:IPUBenchmarks2019, gui2019survey}. This design has the key aim of increasing performance while reducing power usage by minimizing the "bits x millimetres" of data movement during computation \cite{Sze:Survey2017}. }.
    
To scale arbitrarily and leverage all available computation in this "hardware jungle", our key hypothesis is that \textbf{we need inference methods which can operate with distributed local processing / storage and message passing communication}, without the need for a global view or coordination of the whole model \footnote{The opposite proposition would be a global centralized inference method, for example via a monolithic cloud-based processor. This centralized approach would not leverage most of the available compute resources and even if feasible may not be desirable, for communication bandwidth or privacy reasons.}.

\subsection{Gaussian Belief Propagation}

Fortunately, a well-known inference algorithm exists which has these properties: Loopy Belief Propagation \cite{Pearl:book1988, kschischang2001factor, Murphy:etal:1999}.
Belief Progagation (BP) was invented in the 1980s \cite{Pearl:book1988}, and is an algorithm for calculating the marginals of a joint distribution via local message passing between nodes in a factor graph. Factor graphs are a type of bipartite graphical model that connects variables via factors which represent independent relationships \cite{dellaert2017factor, dellaert4factor}.
    
BP guarantees exact marginal computation with one sequence of forward-backward message passing in tree-structured graphs \cite{Bishop:Book2006}, but empirically produces good results when applied to ``loopy" graphs with cycles \cite{Murphy:etal:1999}. 
Unlike Graph neural networks \cite{scarselli2008graph, bronstein2017geometric, battaglia2018relational} which learn edge and node updates that are applied over a fixed number of message passing steps, loopy BP applies probabilistic message passing updates with iterative convergent behaviour.
Although Loopy BP has been successful in a few applications such as error-correcting codes \cite{mceliece1998turbo}, it has not as of yet been applied to broader machine learning problems. 

One issue that has precluded the use of general Loopy BP is that it lacks convergence guarantees.
However, a second and perhaps more relevant issue given modern hardware trends, is that its computational properties have not fitted the dominant processing paradigms of recent decades, CPUs and GPUs. 
Consequently, other factor graph inference algorithms have been preferred which take advantage of global problem structure to operate much more rapidly and robustly than loopy BP on a CPU.

We believe that now is the right time to re-evaluate the properties of loopy BP given the rise of graph-like computing hardware and sparse models.
Indeed there has been a recent resurgence in interest in the algorithm \cite{george2017generative, satorras2021neural, kuck2020belief, Ortiz:etal:CVPR2020, opipari2021differentiable}; in particular, one notable work \cite{Ortiz:etal:CVPR2020} demonstrated that BP on novel graph processor hardware can achieve a 24x speed improvement over standard methods for bundle adjustment.
We focus on the special case of Gaussian Belief Propagation (GBP) \cite{Davison:Ortiz:ARXIV2019}, in which all factors are Gaussian functions of their dependent variables, and we also infer Gaussian-distributed marginals. 
GBP has both more extensive mathematical guarantees \cite{du2018convergence} and stronger empirical performance \cite{Bickson:PhDThesis:2008} than general loopy BP.
It is also simple to implement, with all messages between nodes taking the form of usually low-dimensional Gaussians, which can be represented by small vectors and matrices. 

In this article, we give an introduction to Gaussian Belief Propagation as a strong general purpose algorithmic and representational framework for large scale distributed inference.
Throughout, we present a range of examples across 1D and 2D geometric estimation and image processing in the form of interactive simulations which allow the reader to experiment with and understand the properties of GBP.
We hope that these simulations emphasize the key properties of GBP which we summarize below.

\begin{figure}[h!]
    \centering
    \includegraphics[width=\linewidth]{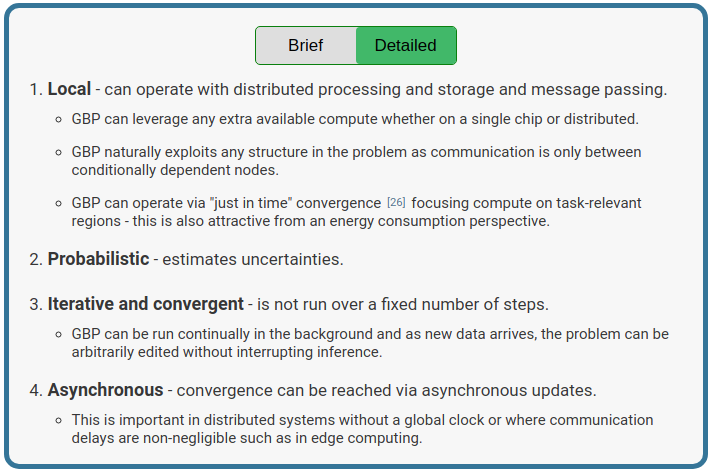}
    \caption{\textcolor{red}{INTERACTIVE FIGURE -- view at \url{https://gaussianbp.github.io}.}}
    \label{fig:key_properties}
\end{figure}

In the remainder of the article, we first introduce the GBP algorithm and show that it is intimately linked to solving a linear system of equations $Ax = b$. 
We then explore 4 key practical details for applying GBP to real problems: extending GBP to handle \hyperref[sec:non-linear-relationships]{non-linear relationships} and \hyperref[sec:non-gaussian-data-distributions]{non-Gaussian data distributions}, using \hyperref[sec:local-updates-and-scheduling]{local message schedules} and lastly how \hyperref[sec:multiscale-learning]{hierarchical structure} can help convergence.

\section{Technical Introduction}

\subsection{Probabilistic Inference}

Inference is the problem of estimating statistical properties of unknown variables $X$ from known or observed quantities $D$ (the data).
For example, one might be interested in inferring tomorrow's weather (X) from historic data (D), or the 3D structure of an environment (X) from a video sequence (D).

Bayesian inference proceeds by first defining a probabilistic model $p(X, D)$ that describes the relationships between data and variables, and then using the sum and product rules of probability
\footnote{
The sum rule is $p(X) = \sum_Y p(X, Y)$ and the product rule is $p(X, Y) = p(Y \rvert X) p(X)$. 
}
to form the posterior distribution $p(X \rvert D) = \frac{p(X, D)}{p(D)}$. The posterior summarizes our belief about $X$ after seeing $D$ and can be used for decision making or other downstream tasks. 
Given the posterior, we can compute various properties of $X$, for example:
\begin{enumerate}
    \item The most likely configuration of the variables $X_{\text{MAP}} = \text{arg max}_X p(X \rvert D)$, or
    \item The marginal posteriors $p(x_i \rvert D) = \sum_{X \setminus x_i} p(X \rvert D)$, which summarize our belief about each individual variable given $D$. 
\end{enumerate}

These two calculations are known as \textbf{maximum a posteriori (MAP) inference} and \textbf{marginal inference} respectively.
An important difference is that MAP inference produces a point estimate while marginal inference retains information about uncertainty.

\subsection{Factor Graphs}

The \href{https://en.wikipedia.org/wiki/Hammersley\%E2\%80\%93Clifford_theorem}{Hammersley-Clifford theorem} tells us that any positive joint distribution $p(X)$ can be represented as a product of factors $f_i$, one per clique, where a clique is a subset of variables $X_i$ in which each variable is connected to all others: 

$$
  p(X) = \prod_i f_i(X_i)
  ~.
$$

\begin{figure}[!h]
    \centering
    \includegraphics[width=0.5\linewidth]{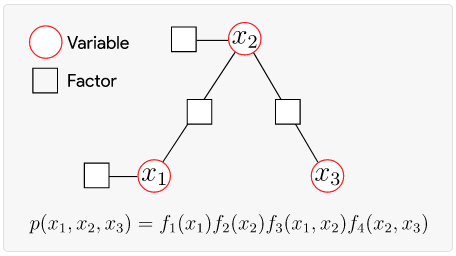}
    \caption{
        \textcolor{red}{INTERACTIVE FIGURE -- view at \url{https://gaussianbp.github.io}.}
        Hover over the factors in the equation to highlight them in the factor graph.
    }
    \label{fig:factorgraph}
\end{figure}

Factorized representations can be very convenient as they expose structure in a model.
\textbf{Factor graphs} are the natural visual representation of this factorization and can be very useful for reasoning about problems \cite{dellaert2017factor}.

In the factor graphs in this article, circles and squares represent variable and factor nodes respectively, with edges connecting each factor to the variables it depends on.
An example of a simple factor graph is shown in Figure \ref{fig:factorgraph}.
By explicitly representing the factors as nodes in the graph, factor graphs clearly emphasize the conditional independence structure of the problem
- the lack of a factor directly connecting two variables means they are conditionally independent 
\footnote{
Mathematically, two variables $x_i$ and $x_j$ are conditionally independent given all other variables $X_{-ij}$ if:
$$
  p(x_i, x_j | X_{-ij}) = p(x_i | X_{-ij}) p(x_j | X_{-ij})
  ~.
$$
An equivalent way condition is: 
$$
  p(x_i | x_j, X_{-ij}) = p(x_i | X_{-ij}) 
  ~.
$$
Intuitively, if $X_{-ij}$ causes both $x_i$ and $x_j$, then if we know $X_{-ij}$ we don't need to know about $x_i$ to predict $x_j$ or about $x_j$ to predict $x_i$.
Conditional independence is often written in shorthand as: $x_i \bot x_j | X_{-ij}$.
}. 

Factor graphs can also be presented as energy based models \cite{lecun2006:EBM} where each factor $f_i$ defines an energy $E_i \geq 0$ associated with a subset of the variables $X_i$ 
\footnote{
This formalism is closely related to the Boltzmann distribution in statistical physics which gives the probability of a state $i$ as a function of the energy of the state and the temperature of the system:
$$
p_i = \frac{e^{-E_i / k T}}{ \sum_j e^{-E_j / k T}}
  ~,
$$
where $k$ is the Boltzmann constant, $T$ is the temperature of the system and $j$ sums over all available states.
}:

$$
f_i(X_i)
\propto
e^{ - E_i(X_i)}
~.
$$

In energy based models, finding the most likely variable configuration is equivalent to minimizing the negative log probability or the sum of factor energies:
$$
    X_{\text{MAP}} = \text{arg min}_X - \log p(X) = \text{arg min}_X \sum_i E_i(X_i)
    ~.
$$
    
\subsection{The Belief Propagation Algorithm}

Belief propagation (BP) is an algorithm for marginal inference, i.e. it computes the marginal posterior distribution for each variable from the set of factors that make up the joint posterior.
BP is intimately linked to factor graphs by the following property: \textbf{BP can be implemented as iterative message passing on the posterior factor graph}. 
The algorithm operates by iteratively updating a node's locally stored belief by sending and receiving messages from neighbouring nodes. Each iteration consists of 3 phases: 

\begin{figure}[h!]
    \centering
    \includegraphics[width=\linewidth]{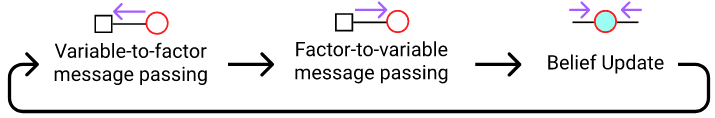}
    \caption{}
    \label{fig:phases}
\end{figure}

\textbf{Belief Update:} The variable node beliefs are updated by taking a product of the incoming messages from all adjacent factors, each of which represents that factor's belief on the receiving node's variables.
\newline
\textbf{Factor-to-variable message:} To send a message to an adjacent variable node, a factor aggregates messages from all other adjacent variable nodes and marginalizes over all the other nodes' variables to produce a message that expresses the factor's belief over the receiving node's variables. 
\newline
\textbf{Variable-to-factor message:} A variable-to-factor message tells the factor what the belief of the variable would be if the receiving factor node did not exist. This is computed by taking the product of the messages the variable node has received from all other factor nodes.
\newline
These 3 operations fully define the algorithm and their equations are presented below.

\begin{figure}[h!]
    \centering
    \includegraphics[width=0.9\linewidth]{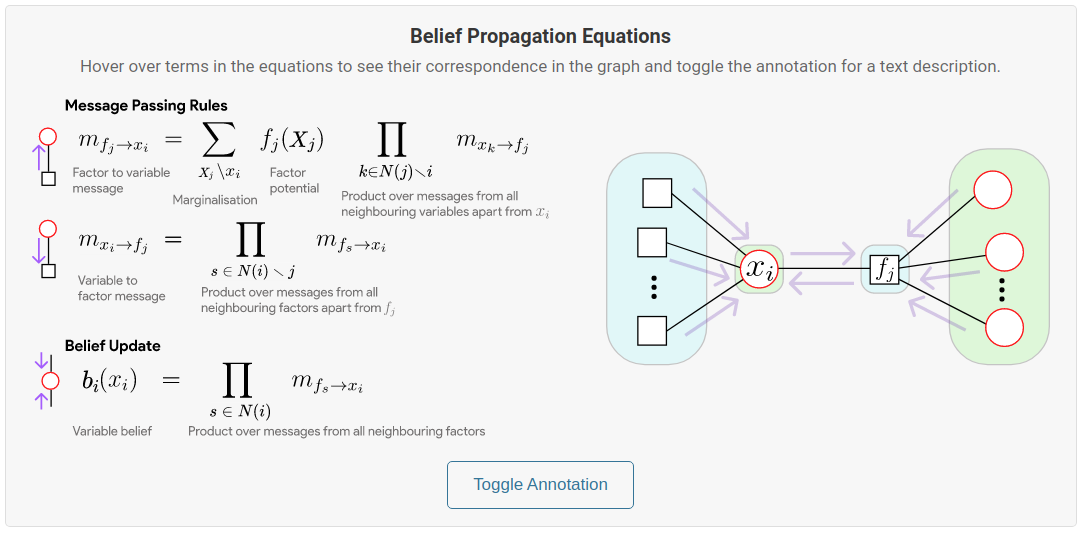}
    \caption{\textcolor{red}{INTERACTIVE FIGURE -- view at \url{https://gaussianbp.github.io}.}}
    \label{fig:bp_equations}
\end{figure}

\begin{figure}[h!]
    \centering
    \includegraphics[width=0.6\linewidth]{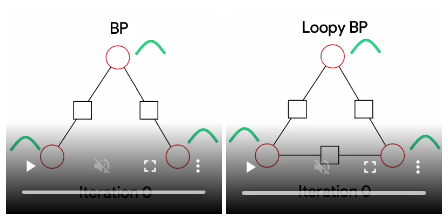}
    \caption{
        \textcolor{red}{INTERACTIVE FIGURE -- view at \url{https://gaussianbp.github.io}.}
        Green curves represent variable belief distributions. \textbf{Left}: BP on a tree. \textbf{Right}:  BP with synchronous updates applied to a graph with a loop. }
    \label{fig:mpvideos}
\end{figure}

Belief propagation was originally developed for graphs that are trees        
\footnote{
  For \href{https://en.wikipedia.org/wiki/Tree_(graph_theory)}{tree} graphs, any two nodes are connected by exactly one path.
}
and the updates were designed such that the beliefs converge to the exact marginals after one sweep of messages from a root node to the leaf nodes and back \cite{Pearl:book1988}. 
For models with arbitrary conditional independence structure, including cycles or "loops", loopy BP \cite{Murphy:etal:1999} iteratively applies the same message passing rules to all nodes.
The simplest variant of loopy BP sends messages from all nodes at every iteration in a synchronous fashion.
The videos in Figure \ref{fig:mpvideos} illustrate how BP is applied to trees and graphs with loops.

As BP was originally developed for trees, its application to loopy graphs was at first empirical \cite{Murphy:etal:1999}. 
Theoretical grounds for applying the same update rules to loopy graphs were later developed \cite{yedidia2000generalized, Weiss:Freeman:NIPS2000, wainwright2008graphical} that explain loopy BP as an approximate variational inference method in which inference is cast as an optimization problem.
Instead of directly minimizing the factor energies (as is done in MAP inference), loopy BP minimizes the \href{https://en.wikipedia.org/wiki/Kullback\%E2\%80\%93Leibler_divergence}{KL divergence} between the posterior and a variational distribution which we use as a proxy for the marginals after optimization.
Loopy BP can be derived via constrained minimization of an approximation of the KL divergence known as the Bethe free energy \cite{yedidia2000generalized} (see \hyperref[sec:derivation]{Appendix A} for the derivation).

As the Bethe free energy is non-convex, loopy BP is not guaranteed to converge and even when it does it may converge to the wrong marginals.
Empirically, however BP generally converges to the true marginals although for very loopy graphs it can fail \cite{Murphy:etal:1999, wainwright2008graphical}. 

Most interesting problems have loopy structures and so for the remainder of the article we will use BP to refer to loopy BP.
So far, although we have outlined the BP equations, we have not specified the form of the factors, messages or beliefs. 
From here, we focus on Gaussian belief propagation which is a special form of continuous BP for Gaussian models. 

\subsection{Gaussian Models}

We are interested in \textbf{Gaussian models} in which all factors and therefore the joint posterior are univariate / multivariate Gaussian distributions. Gaussians are a convenient choice for a number of reasons: (1) they accurately represent the distribution for many real world events \cite{Jaynes:probability2003}, (2) they have a simple analytic form, (3) complex operations can be expressed with simple formulae and (4) they are closed under marginalization, conditioning and taking products (up to normalization). 

\begin{figure}
    \centering
    \includegraphics[width=\linewidth]{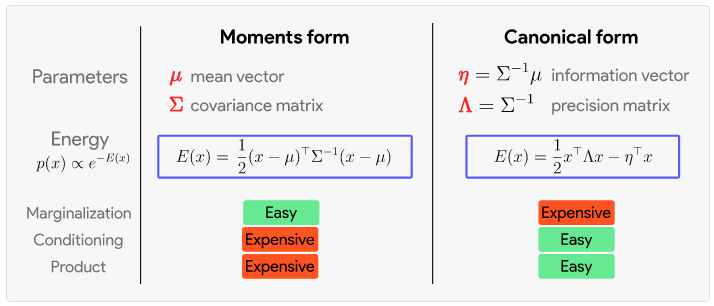}
    \caption{\textcolor{red}{INTERACTIVE FIGURE -- view at \url{https://gaussianbp.github.io}.}}
    \label{fig:gaussianequations}
\end{figure}

A Gaussian factor or in general any Gaussian distribution can be written in the exponential form $p(x) \propto e^{-E(x)}$ with a quadratic energy function. 
There are two ways to write the quadratic energy which correspond to the two common parameterizations of multivariate Gaussian distributions: the \textbf{moments form}\footnote{It's called the moments form as it is parameterized by the first moment and second central moments of the distribution.} and the \textbf{canonical form}. The key properties of each of these parameterizations are summarized in the table in Figure \ref{fig:gaussianequations}. 
  
The canonical form is often preferred when performing inference, for two main reasons. Firstly, taking a product is simple in the canonical form, so it is easy to form the posterior from the factors. Secondly, the precision matrix is sparse and relates closely to the structure of the factor graph. 

\begin{figure}
    \centering
    \includegraphics[width=\linewidth]{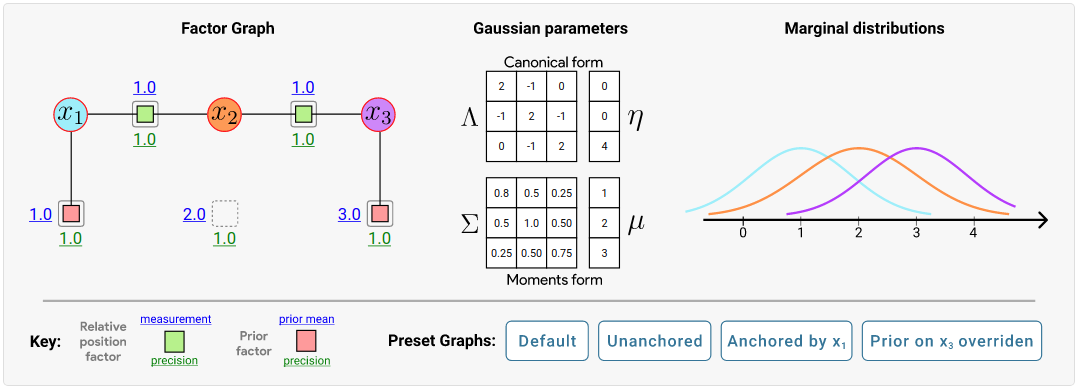}
    \caption{        
    \textcolor{red}{INTERACTIVE FIGURE -- view at \url{https://gaussianbp.github.io}.}
    Click on the factors to turn them on / off and change the factor values by dragging the underlined numbers. You can choose from preset graphs referenced in the text or create your own!
    }
    \label{fig:gaussiangm}
\end{figure}

The precision matrix describes direct associations or conditional dependence between variables. 
If entry $(i,j)$ of the precision matrix is zero then equivalently, there is no factor that directly connects $x_i$ and $x_j$ in the graph. 
You can see this in the default preset graph in Figure \ref{fig:gaussiangm} where $\Lambda_{13}=\Lambda_{31}=0$ and $x_1$ and $x_3$ have no factor directly connecting them.

On the other hand, the covariance matrix describes induced correlations between variables and is dense as long as the graph is one single connected component. 
Unlike the canonical form, the moments form is unable to represent unconstrained distributions, as can be seen by selecting the unanchored preset graph in which there is only relative positional information. 
We encourage the reader to check out the other preset graphs and edit the graph to explore Gaussian models and the relationship with the canonical form.
    
\subsection{From Gaussian Inference to Linear Algebra}
    
The joint distribution corresponding to a factor graph in which all factors are Gaussian can be represented as a single multivariate Gaussian distribution (since the energy terms are additive) in the canonical form:
$$
P(X) \propto \exp( - \frac{1}{2} X^\top \Lambda X + \eta^\top X)
~.
$$
\textbf{MAP inference} corresponds to computing the parameters $X_{\text{MAP}}$ that maximize the above joint distribution. As usual, we can compute the gradient of the log-probability (the total energy):
$$
\nabla_X E = \nabla_X \log P(X) = - \Lambda X + \eta
~,
$$
and solve for $\nabla_X E = 0$. From here, we see that MAP inference in a Gaussian system reduces simply to solving $X_{\text{MAP}} = \Lambda^{-1} \eta = \mu$, which as expected is equal to the mean. 

\textbf{Marginal inference} computes the per-variable marginal posterior distributions.
In the moments form, the marginal distribution of $x_i$ is:
$$
    p(x_i) = \int p(X) dX_{-i}  \propto \exp\big( -\frac{1}{2}(x_i - \mu_i)^\top \Sigma_{ii}^{-1} (x_i - \mu_i) \big)
    ~,
$$
where the mean parameter $\mu_i$ is the $i^{th}$ element of the joint mean vector and the covariance $\Sigma_{ii}$ is entry $(i,i)$ of the joint covariance matrix.
The vector of marginal means for all variables is therefore the joint mean vector $ \mu = \Lambda^{-1} \eta$ = $X_{\text{MAP}}$ and the marginal variances the diagonal entries of the joint covariance matrix $\Sigma = \Lambda^{-1}$.
  
We can therefore summarize inference in Gaussian models as solving the linear system of equations $Ax=b \Leftrightarrow
\Lambda \mu = \eta$. 
\textbf{MAP inference} solves for $\mu$ while \textbf{marginal inference} solves for both  $\mu$ and the block diagonal elements of $\Lambda^{-1}$.

\subsection{Gaussian Belief Propagation}

Having introduced Gaussian models, we now discuss \textbf{Gaussian Belief Propagation (GBP)} a form of BP applied to Gaussian models.
Due to the closure properties of Gaussians, the beliefs and messages are also Gaussians and GBP operates by storing and passing around information vectors and precision matrices.

Unlike general loopy BP, GBP is guaranteed to compute the exact marginal means on convergence, although the same is unfortunately not true for the variances which often converge to the true marginal variances, but are sometimes overconfident for very loopy graphs \cite{Weiss:Freeman:NIPS2000}. 
Although GBP does not in general have convergence guarantees, there some convergence conditions \cite{Bickson:PhDThesis:2008, du2018convergence, su2015convergence} as well as methods to improve chances of convergence (see chapter 22 in \cite{murphy2012machine}). 

\begin{figure}[h!]
    \centering
    \includegraphics[width=0.75\linewidth]{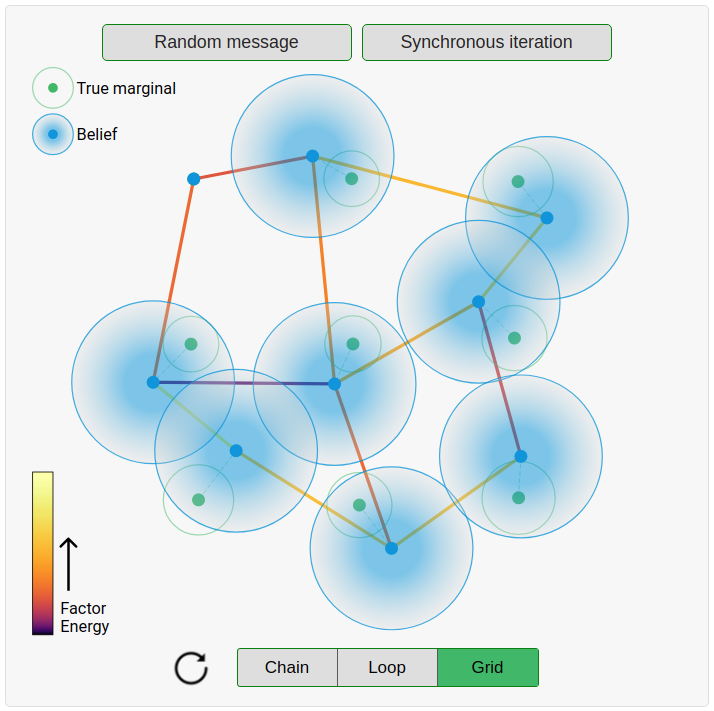}
    \caption{
    \textcolor{red}{INTERACTIVE FIGURE -- view at \url{https://gaussianbp.github.io}.}\newline
    Send messages by clicking on a variable node or by pressing the random or synchronous message buttons at the top. Choose from 3 different graphs with increasing ``loopiness": chain, loop and grid. All problems are anchored by a variable node in the top left with a strong prior that sets the absolute position - this is necessary as all other factors only constrain relative position. All factors are binary constraints (i.e. they connect two variable nodes) and we do not draw the factor nodes but instead draw an edge connecting the two variable nodes whose colour represents the factor energy. We employ message damping for the grid.
    }
    \label{fig:gbpintuition}
\end{figure}

The interactive Figure \ref{fig:gbpintuition} aims to build intuition for GBP by exploring the effect of individual messages.
For easy visualization and interpretation of the beliefs, we examine 3 spatial estimation problems with increasing ``loopiness": a chain, a loop and a grid.
Click on a variable node to send messages to its adjacent variables and observe how neighbouring beliefs are updated.
You will see that GBP converges to the true marginals regardless of the order in which messages are passed
\footnote{    
    Message damping is commonly used to speed up and improve chances of convergence in very loopy graphs, such as the grid problem. 
    Message damping both empirically \cite{Murphy:etal:1999} and theoretically \cite{su2015convergence} improves convergence without affecting the fixed points of GBP.
    The idea behind message damping is to use momentum to reduce chances of oscillation by replacing the message at time $t$ with a combination of the message at time $t$ and time $t-1$:
    $$
        \tilde{m}_{t} = m_{t}^\beta \tilde{m}_{t-1}^{(1 - \beta)}
        ~,
    $$
    which is a weighted sum in log-space:
    $$
        \log \tilde{m}_{t} = \beta \, \log m_{t} + (1 - \beta) \, \log \tilde{m}_{t-1}
        ~.
    $$
    Standard BP is recovered when the damping parameter $\beta = 1$ and $\beta = 0$ corresponds to not updating the message and sending the message from the previous iteration.
    Message damping can be applied to both the variable-to-factor messages and factor-to-variable messages, however we find that applying it just to factor-to-variable messages is sufficient.
    For GBP, message damping corresponds to damping the information vector and precision matrix as a weighted sum: 
    \begin{equation}
        \tilde{\eta}_{t} = \beta \, \eta_{t} + (1 - \beta) \, \tilde{\eta}_{t-1} 
        \;\;\; \text{and} \;\;\;
        \tilde{\Lambda}_{t} = \beta \, \Lambda_{t} + (1 - \beta) \, \tilde{\Lambda}_{t-1} 
        ~.
    \end{equation}
}.

\section{Beyond the standard algorithm}

We have introduced Gaussian Belief Propagation in its basic form as a probabilistic inference algorithm for Gaussian estimation problems. However, to solve real practical problems with GBP, we often need a number of extra details and tricks which we discuss in this section.

\subsection{Non-Gaussian factors}

Although requiring all factors to be Gaussian is a convenient assumption, most interesting problems involve \textbf{non-linear relationships} and / or \textbf{non-Gaussian data distributions}, both of which result in non-Gaussian factors. 
GBP can be extended to handle these problems by linearizing the non-linear relationships and using covariance scaling to handle non-Gaussian data distributions. 

\subsubsection{Non-linear Relationships \label{sec:non-linear-relationships}}

A factor is usually created given some observed data $d$ that we model as $d \sim h(X) + \epsilon$, where $h$ simulates the data generation process from the subset of variables $X$ \footnote{We are overloading $X$ here by using it to denote a subset of the variables for ease of notation.} and $\epsilon \sim \mathcal{N}(0, \Sigma_n)$ is Gaussian noise.
Rearranging, we see that the residual is Gaussian distributed $r = d - h(X) \sim \mathcal{N}(0, \Sigma_n)$, allowing us to form the factor with energy:
$$
E(X) = \frac{1}{2}(h(X) - d)^\top \Sigma_n^{-1} (h(X) - d)
~.
$$

For linear functions $h(X) = \mathtt{J} X + c$, the energy is quadratic in $X$ and we can rearrange the energy so that the factor is in the Gaussian canonical form as:
$$
E(X) = \frac{1}{2} X^\top \Lambda X - \eta^\top X
\;\;\;\;
\text{, where} \;
\eta = \mathtt{J}^\top \Sigma_n^{-1} (d - c) 
\; \text{and} \;
\Lambda = \mathtt{J}^\top \Sigma_n^{-1} \mathtt{J}
~.
$$

If $h$ is non-linear \footnote{The function $h$ could be any non-linear function, for example a trained neural network \cite{czarnowski2020deepfactors} or a Gaussian process \cite{mukadam2018continuous}.}, the energy is no longer quadratic in $X$ meaning the factor is not Gaussian-distributed. 
To restore the Gaussian form, it is standard to use a first-order Taylor expansion about the current estimate $X_{0}$ to approximate the factor as a Gaussian:
$$
h(X) \approx h(X_{0}) + \mathtt{J} (X - X_{0})
~.
$$
Here $\mathtt{J}$ is now the Jacobian matrix and the factor can be written in the same form as above but with $c = h(X_{0}) - \mathtt{J} X_{0}$ \cite{Davison:Ortiz:ARXIV2019}.

After linearization, the posterior is a Gaussian approximation of the true posterior and inference is performed by successively solving linearized versions of the underlying non-linear problem (as in non-linear least squares optimization).

\begin{figure}[h!]
    \centering
    \includegraphics[width=\linewidth]{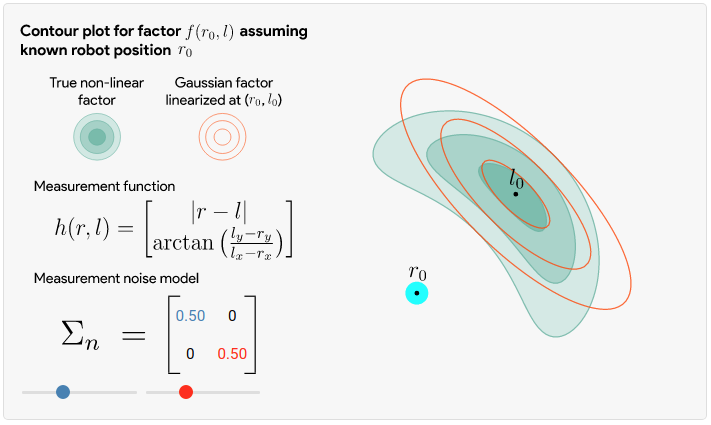}
    \caption{\textcolor{red}{INTERACTIVE FIGURE -- view at \url{https://gaussianbp.github.io}.} Move the two sliders to change the Gaussian noise model. Units of the covariance values are distance$^2$ and radians$^2$.}
    \label{fig:factorlin}
\end{figure}

To see how this linearization works, consider a robot moving in a plane that measures the 2D distance and angle to a landmark also in the plane, where the current estimates for the position of the robot and landmark are $r_0$ and $l_0$ respectively, and the observed measurement $d = h(r_0, l_0)$. 
In the interactive Figure \ref{fig:factorlin}, we show both the true non-linear factor and the Gaussian approximated factor with $r$ held constant at $r_0$. 

The accuracy of the approximate Gaussian factor depends on the linearity of the function $h$ at the linearization point.
As $h$ reasonably is smooth, the linear approximation is good close to the linearization point $l_0$, while further away, the approximation can degrade.
In practice, during optimization we can avoid this region of poor approximation by relinearizing frequently.
As GBP is local, a just-in-time approach to linearization \cite{Ranganathan:etal:IJCAI2007} can be used in which factors are relinearized individually when the current estimate of the adjacent variables strays significantly from the linearization point.

\subsubsection{Non-Gaussian data distributions \label{sec:non-gaussian-data-distributions}}

\begin{figure}[h!]
    \centering
    \includegraphics[width=0.5\linewidth]{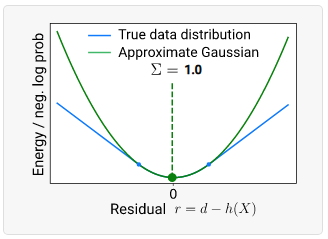}
    \caption{\textcolor{red}{INTERACTIVE FIGURE -- view at \url{https://gaussianbp.github.io}.} Move your mouse over the figure to set the residual and observe how the scaled covariance $\Sigma$ of the approximate Gaussian changes. The true data distribution follows a Huber energy and the blue dots indicate the transition point.}
    \label{fig:scaling}
\end{figure}

A second cause of non-Gaussian factors is non-Gaussian data distributions.
We usually model observed data as coming from a Gaussian distribution: $d \sim h(X) + \epsilon$, by choosing $\epsilon$ to be Gaussian noise.
Although this is generally sensible \cite{Jaynes:probability2003}, true data distributions often have stronger tails or are more tightly peaked.
In these cases, to retain the Gaussian form for GBP, we use an approximate Gaussian data distribution via covariance scaling \cite{Agarwal:etal:ICRA2013, Davison:Ortiz:ARXIV2019}. 
The covariance of the approximate Gaussian is chosen such that the quadratic energy matches the true non-Gaussian energy at that residual, as shown in Figure \ref{fig:scaling}.

Robust data distributions (or \href{https://en.wikipedia.org/wiki/M-estimator}{M-estimators}), such as the Huber energy \cite{Huber:AMS:1964, Huber:1981}, are a common class of non-Gaussian data distributions which have greater probability mass in the tails to reduce sensitivity to outliers.
The Huber energy is a continuous function that is quadratic close to the mean and transitions to a linear function for large residuals to reduce the energy cost of outliers
\footnote{
The distribution induced by the Huber energy is a Gaussian distribution close to the mean and a Laplace distribution in the tails.
The probability density function for the Laplace distribution is:
$$
p(x ; \mu, \beta) = \frac{1}{2b} \exp\big(\frac{-|x - \mu|}{b}\big)
  ~.
$$
}:
$$
E_{\text{huber}}(r) =     
\begin{cases}
  \frac{1}{2} r^\top \Sigma_n^{-1} r, & \text{if}\ \rvert r\rvert  < t \\
  A \;+ \;  B \rvert r \rvert, & \text{otherwise} ~.
\end{cases}
$$
The approximate quadratic energy for the Huber distribution can be found by solving $\frac{1}{2} r^\top \Sigma_{\text{sc}}^{-1} r = E_{\text{huber}}(r)$ to give the diagonal scaled covariance:
$$
\Sigma_{\text{sc}} =
\begin{cases}
  \Sigma_n , & \text{if}\ \rvert r\rvert  < t \\
  \frac{2 E_{\text{huber}}(r)}{ r^\top \Sigma_n^{-1} r} \Sigma_n, & \text{otherwise} ~.
\end{cases}
$$

Robust energy functions can make GBP much more generally useful - for example, they are crucial for bundle adjustment \cite{Ortiz:etal:CVPR2020} and for sharp image denoising (as we will see later in Figure \ref{fig:attentiongl}).
More generally, our interpretation is that robust factors can play a similar role to non-linearities in neural networks, activating or deactivating messages in the graph.

The interactive Figure \ref{fig:gbp1drobust} below gives intuition for the effect of robust factors for the task of 1D line fitting.
The variables we are estimating are the $y$ values at fixed intervals along the $x$ axis and the blue circles and lines show the mean and standard deviation of the beliefs. 
The red squares are measurements that produce data factors in the graph and there are also smoothness factors between all adjacent variable nodes encouraging the $y$ values to be close.
You can add your own data factors by clicking on the canvas and a diagram of the factor graph is in the bottom right of the figure.
Press play to run synchronous GBP and observe that a Huber energy can disregard outliers and retain step discontinuities in the data unlike the standard squared loss.

\begin{figure}[h!]
    \centering
    \includegraphics[width=\linewidth]{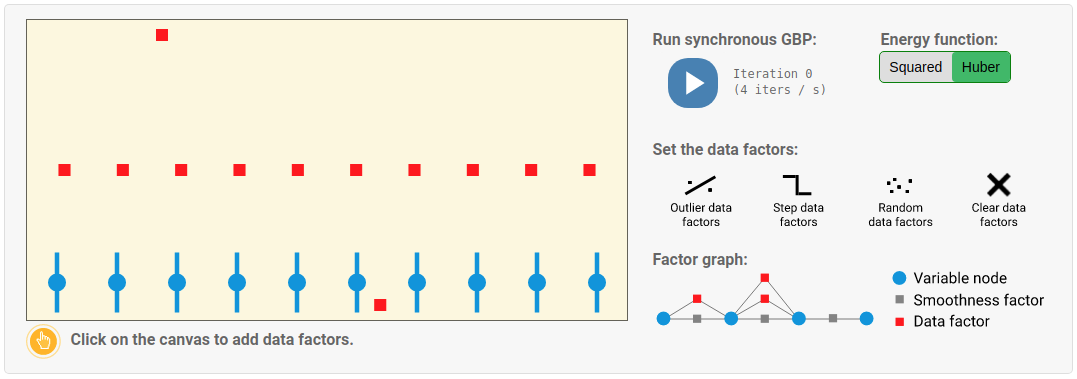}
    \caption{\textcolor{red}{INTERACTIVE FIGURE -- view at \url{https://gaussianbp.github.io}.} 1D line fitting. Create data factors by clicking on the canvas or use a preset data factor configuration with the outlier, step and random buttons. Note the variable nodes start at the bottom of the canvas before they have received any messages.}
    \label{fig:gbp1drobust}
\end{figure}

\subsection{Local updates and Scheduling \label{sec:local-updates-and-scheduling}}
    
So far, we have assumed that all variable and factor nodes broadcast messages at each iteration in a synchronous fashion, where all nodes absorb and broadcast messages in parallel.
In fact, this is far from a requirement and as GBP is entirely local, messages can be sent arbitrarily and asynchronously. 

It turns out that the message schedule can have a dramatic effect on the rate of convergence. For example, swapping synchronous updates for random message passing tends to improve convergence, while a fixed ``round-robin" schedule can do even better \cite{koller2009probabilistic}.
Better yet, if each message requires some unit of computation (and therefore energy), it's possible to prioritize sending messages that we think will contribute most to the overall convergence of the system (there is evidence that the brain may apply a similar economical principle \cite{Evans:Burgess:NIPS2019}).
This is the idea behind residual belief propagation (RBP) \cite{Elidan:etal:UAI2006} and similar variants \cite{Sutton:McCallum:UAI2007, Ranganathan:etal:IJCAI2007}, which form a message queue according to the norm of the difference from the previous message.

\begin{figure}[h!]
    \centering
    \includegraphics[width=\linewidth]{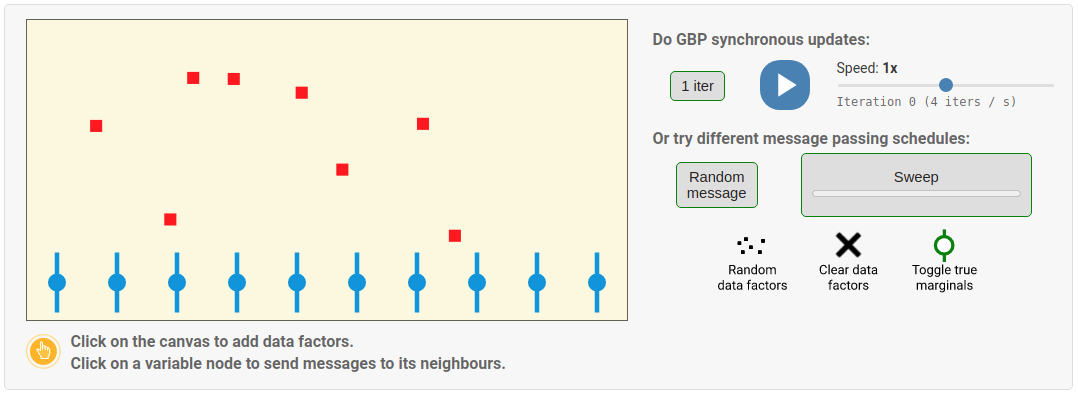}
    \caption{\textcolor{red}{INTERACTIVE FIGURE -- view at \url{https://gaussianbp.github.io}.} 1D line fitting. Experiment with different message passing schedules. Use the synchronous, random and sweep schedules or click on a variable node to send messages to its neighbours. Note the variable nodes start at the bottom of the canvas before they have received any messages.}
    \label{fig:gbp1d}
\end{figure}

In Figure \ref{fig:gbp1d}, we explore message scheduling using the 1D line fitting task once again.
The underlying factor graph is a chain (no loops) and so will converge after one sweep of messages from left to right and back again.
You can send messages through the graph using the preset schedules (synchronous, random or sweep) or create your own schedule by clicking on a variable node to send messages outwards.

Playing around with different schedules for surface estimation highlights two important properties of GBP.
First, GBP can converge with an arbitrary message passing schedule. 
As a consequence, GBP can readily operate in systems with no global clock and varying local compute budgets such as on neuromorphic hardware or between a group of distributed devices \cite{micusik2020ego}.

The second property is that GBP can achieve approximate local convergence without global convergence. 
Due to the factorized structure of GBP \cite{diehl2018factorized}, global inference is achieved by jointly solving many interdependent local subproblems.
There are many instances in which we might only be interested in local solutions - in these cases, GBP can operate in a \textbf{just-in-time} or \textbf{attention-driven} fashion, focusing processing on parts of the graph to solve local subproblems as the task demands. 
Local message passing can yield accurate relative local solutions which estimate the marginals up to global corrections that come from more distant parts of the graph \footnote{One simple example is mapping two connected rooms. An accurate local map of one room can be constructed by focusing processing on the part of the factor graph in that room. For some applications this may be sufficient while for others it may be important to build a map with an accurate absolute position which may require longer range message passing between the parts of the graph corresponding to each separate room. }.
This attention-driven scheduling can be very economical with compute and energy, only sending the most task-critical messages. 

\begin{figure}[h!]
    \centering
    \includegraphics[width=\linewidth]{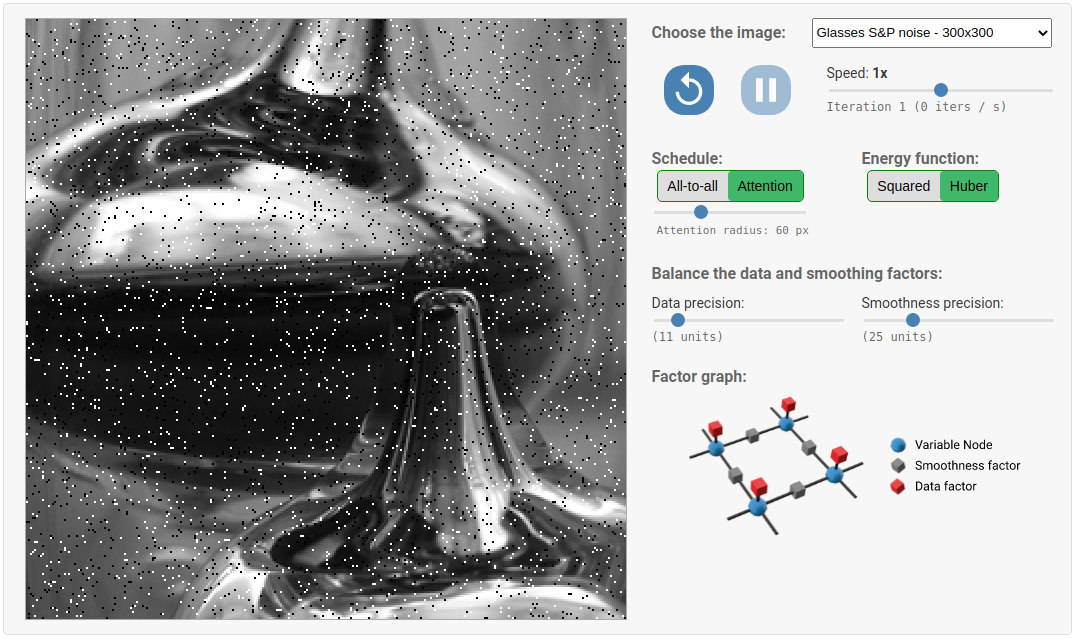}
    \caption{\textcolor{red}{INTERACTIVE FIGURE -- view at \url{https://gaussianbp.github.io}.} GBP for image denoising \cite{Scona:github2021}. You can choose either all-to-all message passing or an attention-driven schedule around the mouse. Choosing a squared loss function results in a blurred denoising while the robust Huber loss gives a sharper denoising. You can select from a number of preloaded images with salt and pepper (S\&P) or Gaussian noise or use a live feed from your webcam.}
    \label{fig:attentiongl}
\end{figure}

In Figure \ref{fig:attentiongl} below we explore attention-driven message passing for image denoising \footnote{As there are no long-range connections in the image denoising graph, local message passing can produce the true local marginals as the effect of more distant parts of the graph is negligible.}. 
Image denoising is the 2D equivalent of the surface estimation problem from the Figure \ref{fig:gbp1d}. 
The only difference is that previously although variable nodes were at discrete locations the data factors were at any location, while now the data factors are at the same discrete locations as the variable nodes with one per node.
We also revisit the use of robust energy functions with GBP via covariance scaling which is crucial for sharp denoising. 

\subsection{Multiscale Learning \label{sec:multiscale-learning}}

Propagating information from one node to another with GBP takes the same number of iterations as the number of hops between the nodes. 
For nearby nodes in a local region, information can be communicated in a small number of iterations and consensus can be reached quickly, while for distant nodes, a global consensus can take many more iterations to be established. 
This is an inherent property of local algorithms and can be summarized as low frequency errors decay more slowly than the high frequency errors.

Regular grid structured graphs appear a lot in computer vision (e.g. image segmentation) and in discretized boundary value problems (e.g. solving for the temperature profile along a rod).
Accelerating convergence in such grid graphs has been well-studied in the field of Multigrid methods \cite{briggs2000multigrid}.
One simple approach is to coarsen the grid which transforms low frequency errors into higher frequency errors that decay faster. 
After convergence in the coarsened grid, the solution is used to initialize inference in the original grid which now has smaller low frequency errors. 
This is the idea behind coarse-to-fine optimization which is used in many grid-based problems where it is simple to build a coarser graph. In one notable work \cite{felzenszwalb2006efficient}, the authors demonstrate much faster inference for stereo, optical flow and image restoration with multiscale BP.

Mulitgrid methods can only be applied to graphs with a grid-like structure where it is possible to build equivalent coarsened representations. 
In general, most problems are more unstructured and it is not clear how to build a coarsened or abstracted representation of the original problem.
In the general case, we see two possible ways to build hierarchy into a model. A network could be trained to directly predict specific abstractions that form long range connections when included in the graph. Second, the graph could contain additional constraints that define a generic hierarchical structure (much like a neural network) and then the abstractions themselves are also inferred \cite{george2017generative}.

\section{Related Methods}

Solving real non-linear problems with GBP is done by iteratively solving linearized Gaussian versions of the true non-linear problem. 
This general pattern of successively solving linearized problems underpins many different non-linear inference methods.
There are efficient libraries \cite{CeresManual, Dellaert:TechReport2012, kummerle2011g20} for non-linear inference which use trust region methods like Gauss-Newton or line search to guide the repeated linear steps \footnote{Trust region methods approximate the energy using a model within a trust region. For example, the Gauss-Newton method uses a quadratic model meaning the factors are approximated as Gaussians as in GBP. Line search methods choose a descent direction and then step size at each iteration. In trust region methods, the most expensive step is solving the linear system, while for line search methods choosing the direction is the most expensive part.}.

GBP is just one of many possible algorithms that can be used to solve the linearized Gaussian model.
To place GBP amongst other methods, we present an overview of a number of related methods for MAP and marginal inference for Gaussian models in the \hyperref[fig:related]{table} below. 
As a reminder, inference in Gaussian models is equivalent to solving the linear system $\Lambda \mu = \eta$, for $\mu$ in MAP inference and for $\mu$ and the diagonal elements of $\Lambda^{-1}$ in marginal inference.
You can hover over the circles in the figure to explore how GBP relates to other methods. 

\begin{figure}
    \centering
    \includegraphics[width=\linewidth]{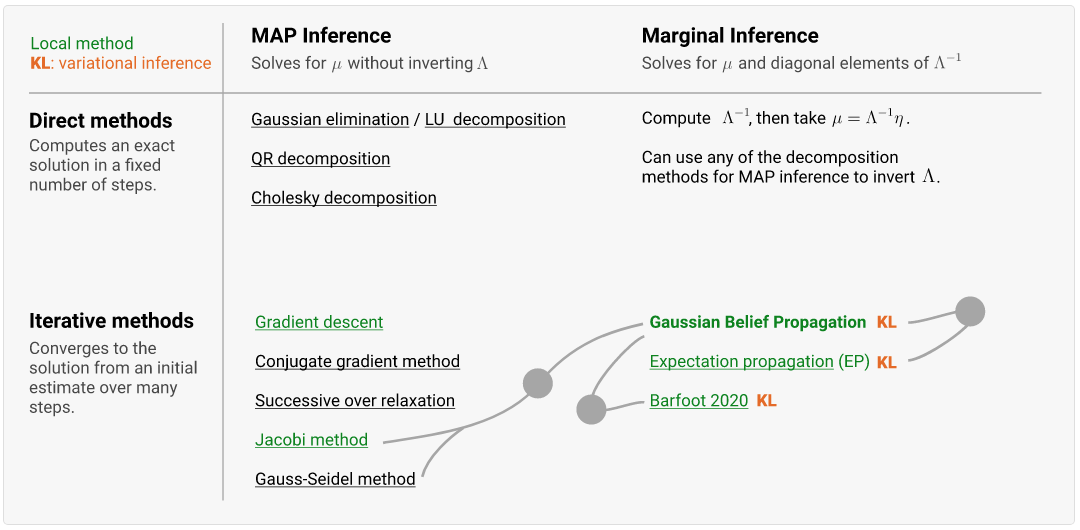}
    \caption{
    \textcolor{red}{INTERACTIVE FIGURE -- view at \url{https://gaussianbp.github.io}.}
    Hover over the circles to show comments. Green methods are local, by which we mean that they can be implemented as a distributed message passing algorithm in which all nodes compute messages on every iteration and each node communicates only with a small local neighbourhood. That does not mean to say that the other methods are entirely centralised; many of the methods are somewhat parallel or have modified versions that are more strongly parallel (e.g. parallel conjugate gradient). The KL symbol indicates a variational inference method. Note that this is not a comprehensive list - for example we have excluded sampling based inference methods. [References: Expectation propagation \cite{minka2013ep}, Barfoot 2020 \cite{barfoot2020fundamental}]}
    \label{fig:related}
\end{figure}

With so many different inference methods, choosing which method to use can be a challenge in itself. 
Judging the speed of each method is complex and depends on both the sparsity structure of $\Lambda$ and on the implementation on the available hardware. 
Our key argument in this article is that we want a general method that is local, probabilistic and iterative which led us towards Gaussian Belief Propagation.

Other notable candidate methods that are local, probabilistic and iterative are Expectation Propagation (EP) \cite{minka2013ep} and Barfoot's algorithm \cite{barfoot2020fundamental}.
EP is generally not node-wise parallel and simplifies to GBP in the special case when it is node-wise parallel, while Barfoot's algorithm involves extra communication edges and is yet to be applied to real problems. 
For these reasons GBP stands out as the extreme case that maximizes parallelism and minimizes communication - two principles that are at the core of scalable and low-power computation.

\section{Conclusions}

We envisage that ML systems of the future will be large scale, heterogeneous and distributed and as such will require flexible and scalable probabilistic inference algorithms.
In this article, we argued that Gaussian Belief Propagation is a strong candidate algorithm as it is local, probabilistic, iterative and asynchronous.
Additionally, we showed 1) how GBP is much more general with a prescription for handling non-linear factors and robust energy functions, 2) how GBP can operate in an attention-driven fashion and 3) how hierarchical structure can help convergence.
We hope that this visual introduction will enourage more researchers and practitioners to look into GBP as an alternative to existing inference algorithms.

We see many exciting directions for future research around GBP and provide a \href{https://colab.research.google.com/drive/1-nrE95X4UC9FBLR0-cTnsIP_XhA_PZKW?usp=sharing}{GBP Library Colab Notebook} as a starting point for the interested reader. 
Some directions we are most excited about are improving theoretical guarantees, using learned factors \cite{czarnowski2020deepfactors, mukadam2018continuous, opipari2021differentiable}, introducing discrete variables, combining GBP with GNNs \cite{satorras2021neural, kuck2020belief}, incrementally abstracting factor graphs, investigating numerical precision for messages, using GBP for distributed learning in overparameterized networks and lastly unifying iterative inference with test-time self-supervised learning.

\section{Supplementary Playground}

\begin{figure}
    \centering
    \includegraphics[width=\linewidth]{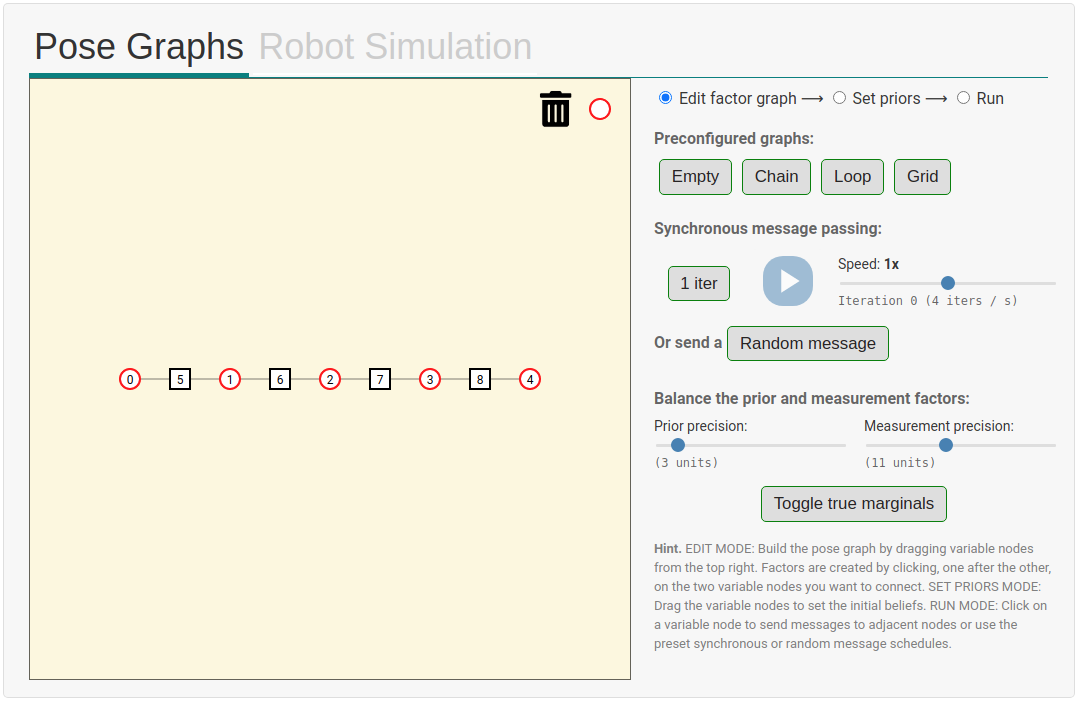}
    \caption{\textcolor{red}{INTERACTIVE FIGURE -- view at \url{https://gaussianbp.github.io}.}}
    \label{fig:playground}
\end{figure}

We encourage the interested reader to explore our supplementary \hyperref[fig:playground]{GBP playground}.
The playground consists of 2 interative diagrams based around 2D pose graphs.
In the first you can construct your own pose graph, set the initialization and then choose how messages are passed through the graph.
The second is a simulation of a robot exploring a 2D environment with landmarks. 

\section{Acknowledgments}
      We are grateful to many researchers with whom we have discussed some of the ideas in this paper, especially from the Dyson Robotics Lab and Robot Vision Group at Imperial College London.
      We would particularly like to thank Xiaofan Mu, Raluca Scona, Riku Murai, Edgar Sucar, Seth Nabarro, Tristan Laidlow, Nanfeng Liu, Shuaifeng Zhi, Kentara Wada and Stefan Leutenegger.

{\small
\bibliographystyle{abbrv}
\bibliography{bibliography}

\begin{thebibliography}{10}

\bibitem{Agarwal:etal:ICRA2013}
P.~Agarwal, G.~D. Tipaldi, L.~Spinello, C.~Stachniss, and W.~Burgard.
\newblock Robust map optimization using dynamic covariance scaling.
\newblock In {\em Proceedings of the {IEEE} International Conference on
  Robotics and Automation ({ICRA})}, 2012.

\bibitem{CeresManual}
S.~Agarwal, M.~K., and Others.
\newblock Ceres solver.
\newblock \url{http://ceres-solver.org}.

\bibitem{barfoot2020fundamental}
T.~D. Barfoot.
\newblock Fundamental linear algebra problem of gaussian inference.
\newblock {\em arXiv preprint arXiv:2010.08022}, 2020.

\bibitem{battaglia2018relational}
P.~W. Battaglia, J.~B. Hamrick, V.~Bapst, A.~Sanchez-Gonzalez, V.~Zambaldi,
  M.~Malinowski, A.~Tacchetti, D.~Raposo, A.~Santoro, R.~Faulkner, et~al.
\newblock Relational inductive biases, deep learning, and graph networks.
\newblock {\em arXiv preprint arXiv:1806.01261}, 2018.

\bibitem{bengio2017consciousness}
Y.~Bengio.
\newblock The consciousness prior.
\newblock {\em arXiv preprint arXiv:1709.08568}, 2017.

\bibitem{Bickson:PhDThesis:2008}
D.~Bickson.
\newblock {\em Gaussian belief propagation: Theory and Application}.
\newblock PhD thesis, PhD thesis, The Hebrew University of Jerusalem, 2008.

\bibitem{Bishop:Book2006}
C.~M. Bishop.
\newblock {\em Pattern Recognition and Machine Learning}.
\newblock Springer-Verlag New York, Inc., 2006.

\bibitem{briggs2000multigrid}
W.~L. Briggs, V.~E. Henson, and S.~F. McCormick.
\newblock {\em A multigrid tutorial}.
\newblock SIAM, 2000.

\bibitem{bronstein2017geometric}
M.~M. Bronstein, J.~Bruna, Y.~LeCun, A.~Szlam, and P.~Vandergheynst.
\newblock Geometric deep learning: going beyond euclidean data.
\newblock {\em IEEE Signal Processing Magazine}, 34(4):18--42, 2017.

\bibitem{czarnowski2020deepfactors}
J.~Czarnowski, T.~Laidlow, R.~Clark, and A.~J. Davison.
\newblock Deepfactors: Real-time probabilistic dense monocular slam.
\newblock {\em IEEE Robotics and Automation Letters}, 5(2):721--728, 2020.

\bibitem{davison2018futuremapping}
A.~J. Davison.
\newblock Futuremapping: The computational structure of spatial ai systems.
\newblock {\em arXiv preprint arXiv:1803.11288}, 2018.

\bibitem{Davison:Ortiz:ARXIV2019}
A.~J. Davison and J.~Ortiz.
\newblock Futuremapping 2: Gaussian belief propagation for spatial ai.
\newblock {\em arXiv preprint arXiv:arXiv:1910.14139}, 2019.

\bibitem{Dellaert:TechReport2012}
F.~Dellaert.
\newblock Factor graphs and gtsam.
\newblock Technical Report GT-RIM-CP\&R-2012-002, Georgia Institute of
  Technology, 2012.

\bibitem{dellaert4factor}
F.~Dellaert.
\newblock Factor graphs: Exploiting structure in robotics.
\newblock {\em Annual Review of Control, Robotics, and Autonomous Systems}, 4,
  2021.

\bibitem{dellaert2017factor}
F.~Dellaert and M.~Kaess.
\newblock Factor graphs for robot perception.
\newblock {\em Foundations and Trends in Robotics}, 6(1-2):1--139, 2017.

\bibitem{diehl2018factorized}
P.~U. Diehl, J.~Martel, J.~Buhmann, and M.~Cook.
\newblock Factorized computation: What the neocortex can tell us about the
  future of computing.
\newblock {\em Frontiers in computational neuroscience}, 12:54, 2018.

\bibitem{du2018convergence}
J.~Du, S.~Ma, Y.-C. Wu, S.~Kar, and J.~M. Moura.
\newblock Convergence analysis of belief propagation on gaussian graphical
  models.
\newblock {\em arXiv preprint arXiv:1801.06430}, 2018.

\bibitem{Elidan:etal:UAI2006}
G.~Elidan, I.~McGraw, and D.~Koller.
\newblock Residual belief propagation: Informed scheduling for asynchronous
  message passing.
\newblock In {\em In Proceedings of the Conference on Uncertainty in Artificial
  Intelligence (UAI)}, 2006.

\bibitem{Evans:Burgess:NIPS2019}
T.~Evans and N.~Burgess.
\newblock Coordinated hippocampal-entorhinal replay as structural inference.
\newblock In {\em Neural Information Processing Systems (NeurIPS)}, volume~32,
  2019.

\bibitem{fedus2021switch}
W.~Fedus, B.~Zoph, and N.~Shazeer.
\newblock Switch transformers: Scaling to trillion parameter models with simple
  and efficient sparsity.
\newblock {\em arXiv preprint arXiv:2101.03961}, 2021.

\bibitem{felzenszwalb2006efficient}
P.~F. Felzenszwalb and D.~P. Huttenlocher.
\newblock Efficient belief propagation for early vision.
\newblock {\em International journal of computer vision}, 70(1):41--54, 2006.

\bibitem{george2017generative}
D.~George, W.~Lehrach, K.~Kansky, M.~L{\'a}zaro-Gredilla, C.~Laan, B.~Marthi,
  X.~Lou, Z.~Meng, Y.~Liu, H.~Wang, et~al.
\newblock A generative vision model that trains with high data efficiency and
  breaks text-based captchas.
\newblock {\em Science}, 358(6368), 2017.

\bibitem{Ghahramani:Nature2015}
Z.~Ghahramani.
\newblock Probabilistic machine learning and artificial intelligence.
\newblock {\em Nature}, 521(7553):452--459, 2015.

\bibitem{gui2019survey}
C.-Y. Gui, L.~Zheng, B.~He, C.~Liu, X.-Y. Chen, X.-F. Liao, and H.~Jin.
\newblock A survey on graph processing accelerators: Challenges and
  opportunities.
\newblock {\em Journal of Computer Science and Technology}, 34(2):339--371,
  2019.

\bibitem{Huber:AMS:1964}
P.~J. Huber.
\newblock Robust estimation of a location parameter.
\newblock {\em The Annals of Mathematical Statistics}, 35(1):pp. 73--101, 1964.

\bibitem{Huber:1981}
P.~J. Huber.
\newblock {\em Robust Statistics}.
\newblock Wiley Series in Probability and Statistics. Wiley-Interscience, 1981.

\bibitem{Jaynes:probability2003}
E.~T. Jaynes.
\newblock {\em Probability theory: The logic of science}.
\newblock Cambridge university press, 2003.

\bibitem{koller2009probabilistic}
D.~Koller and N.~Friedman.
\newblock {\em Probabilistic graphical models: principles and techniques}.
\newblock MIT press, 2009.

\bibitem{kschischang2001factor}
F.~R. Kschischang, B.~J. Frey, and H.-A. Loeliger.
\newblock Factor graphs and the sum-product algorithm.
\newblock {\em IEEE Transactions on information theory}, 47(2):498--519, 2001.

\bibitem{kuck2020belief}
J.~Kuck, S.~Chakraborty, H.~Tang, R.~Luo, J.~Song, A.~Sabharwal, and S.~Ermon.
\newblock Belief propagation neural networks.
\newblock In {\em Neural Information Processing Systems (NeurIPS)}, 2020.

\bibitem{kummerle2011g20}
R.~Kummerle, G.~Grisetti, H.~Strasdat, K.~Konolige, and W.~Burgard.
\newblock g 2 o: A general framework for graph optimization.
\newblock In {\em 2011 IEEE International Conference on Robotics and
  Automation}, pages 3607--3613. IEEE, 2011.

\bibitem{Lacey:IPUBenchmarks2019}
D.~Lacey.
\newblock Graphcore ipu benchmarks.
\newblock URL https://www.graphcore.ai/posts/new-graphcore-ipu-benchmarks,
  2019.

\bibitem{lecun2006:EBM}
Y.~LeCun, S.~Chopra, R.~Hadsell, M.~Ranzato, and F.~Huang.
\newblock A tutorial on energy-based learning.
\newblock {\em Predicting structured data}, 1(0), 2006.

\bibitem{mceliece1998turbo}
R.~J. McEliece, D.~J.~C. MacKay, and J.-F. Cheng.
\newblock Turbo decoding as an instance of pearl's "belief propagation"
  algorithm.
\newblock {\em IEEE Journal on selected areas in communications},
  16(2):140--152, 1998.

\bibitem{micusik2020ego}
B.~Micusik and G.~Evangelidis.
\newblock Ego-motion alignment from face detections for collaborative augmented
  reality.
\newblock {\em arXiv preprint arXiv:2010.02153}, 2020.

\bibitem{minka2013ep}
T.~P. Minka.
\newblock Expectation propagation for approximate bayesian inference.
\newblock {\em arXiv preprint arXiv:1301.2294}, 2013.

\bibitem{mukadam2018continuous}
M.~Mukadam, J.~Dong, X.~Yan, F.~Dellaert, and B.~Boots.
\newblock Continuous-time gaussian process motion planning via probabilistic
  inference.
\newblock {\em The International Journal of Robotics Research},
  37(11):1319--1340, 2018.

\bibitem{murphy2012machine}
K.~P. Murphy.
\newblock {\em Machine learning: a probabilistic perspective}.
\newblock MIT press, 2012.

\bibitem{Murphy:etal:1999}
K.~P. Murphy, Y.~Weiss, and M.~I. Jordan.
\newblock Loopy belief propagation for approximate inference: An empirical
  study.
\newblock In {\em Proceedings of the Fifteenth Conference on Uncertainty in
  Artificial Intelligence}, 1999.

\bibitem{opipari2021differentiable}
A.~Opipari, C.~Chen, S.~Wang, J.~Pavlasek, K.~Desingh, and O.~C. Jenkins.
\newblock Differentiable nonparametric belief propagation.
\newblock {\em arXiv preprint arXiv:2101.05948}, 2021.

\bibitem{Ortiz:etal:CVPR2020}
J.~Ortiz, M.~Pupilli, S.~Leutenegger, and A.~Davison.
\newblock Bundle adjustment on a graph processor.
\newblock In {\em Proceedings of the IEEE Conference on Computer Vision and
  Pattern Recognition (CVPR)}, 2020.

\bibitem{Pearl:book1988}
J.~Pearl.
\newblock {\em Probabilistic reasoning in intelligent systems: networks of
  plausible inference}.
\newblock Morgan Kaufmann, 1988.

\bibitem{Ranganathan:etal:IJCAI2007}
A.~Ranganathan, M.~Kaess, and F.~Dellaert.
\newblock Loopy sam.
\newblock In {\em Proceedings of the International Joint Conference on
  Artificial Intelligence (IJCAI)}, 2007.

\bibitem{satorras2021neural}
V.~G. Satorras and M.~Welling.
\newblock Neural enhanced belief propagation on factor graphs.
\newblock In {\em International Conference on Artificial Intelligence and
  Statistics}, pages 685--693. PMLR, 2021.

\bibitem{scarselli2008graph}
F.~Scarselli, M.~Gori, A.~C. Tsoi, M.~Hagenbuchner, and G.~Monfardini.
\newblock The graph neural network model.
\newblock {\em IEEE transactions on neural networks}, 20(1):61--80, 2008.

\bibitem{Scona:github2021}
R.~Scona.
\newblock Belief propagation denoising project, 2021.

\bibitem{shazeer2017outrageously}
N.~Shazeer, A.~Mirhoseini, K.~Maziarz, A.~Davis, Q.~Le, G.~Hinton, and J.~Dean.
\newblock Outrageously large neural networks: The sparsely-gated
  mixture-of-experts layer.
\newblock In {\em Proceedings of the International Conference on Learning
  Representations (ICLR)}, 2017.

\bibitem{su2015convergence}
Q.~Su and Y.-C. Wu.
\newblock On convergence conditions of gaussian belief propagation.
\newblock {\em IEEE Transactions on Signal Processing}, 63(5):1144--1155, 2015.

\bibitem{Sutter:Jungle2011}
H.~Sutter.
\newblock Welcome to the jungle.
\newblock URL https://herbsutter.com/welcome-to-the-jungle, 2011.

\bibitem{Sutton:McCallum:UAI2007}
C.~Sutton and A.~McCallum.
\newblock Improved dynamic schedules for belief propagation.
\newblock In {\em In Proceedings of the Conference on Uncertainty in Artificial
  Intelligence (UAI)}, 2007.

\bibitem{Sutton:BitterLesson2019}
R.~Sutton.
\newblock The bitter lesson.
\newblock \url{http://www.incompleteideas.net/IncIdeas/BitterLesson.html},
  2019.

\bibitem{Sze:Survey2017}
V.~Sze, Y.-H. Chen, T.-J. Yang, and J.~S. Emer.
\newblock Efficient processing of deep neural networks: A tutorial and survey.
\newblock {\em Proceedings of the IEEE}, 105(12):2295--2329, 2017.

\bibitem{wainwright2008graphical}
M.~J. Wainwright and M.~I. Jordan.
\newblock {\em Graphical models, exponential families, and variational
  inference}.
\newblock Now Publishers Inc, 2008.

\bibitem{Weiss:Freeman:NIPS2000}
Y.~Weiss and W.~T. Freeman.
\newblock Correctness of belief propagation in gaussian graphical models of
  arbitrary topology.
\newblock In {\em Neural Information Processing Systems (NeurIPS)}, 2000.

\bibitem{yedidia2000generalized}
J.~S. Yedidia, W.~T. Freeman, Y.~Weiss, et~al.
\newblock Generalized belief propagation.
\newblock In {\em NIPS}, volume~13, pages 689--695, 2000.

\end{thebibliography}
}

\newpage
\section*{Appendix A: Variational BP Derivation} \label{sec:derivation}

This standard derivation of the belief propagation equations follows Yedida et al \cite{yedidia2000generalized} in showing that the BP update rules follow from constrained minimization of an approximate free energy known as the Bethe free energy. 

We begin by writing the posterior as a product of the factors:
$$
p(X) = \frac{1}{Z} \prod_a f_a(X_a) = \frac{1}{Z} \prod_a e^{-E(X_a)}
~.
$$
The goal of variational inference is to find a variational distribution $q(X)$ that approximates the posterior well by minimizing the Kullback-Leibler divergence between the variational distribution and the posterior.
The KL divergence is a non-negative asymmetric similarity metric that has a minimum of 0 when $p = q$.
$$
\begin{aligned}
  KL(q \lvert \rvert p) &= \sum_{X} q(X) \log \frac{q(X)}{p(X)} \\
    &= \sum_{X} q(X) \log q(X) - \sum_{X} q(X) \log p(X) \\
    &= -H_q(X) - \mathop{\mathbb{E}}_q [\log p(X)] \\
    &= -H_q(X) - \sum_{a} \mathop{\mathbb{E}}_q [\log f_a(X_a)] + \log(Z) \\
    &= F(p, q) + \log(Z)
\end{aligned}
$$
Above, we defined the free energy: 
$$
F(p, q) = -H_q(X) - \sum_{a} \mathop{\mathbb{E}}_q [\log f_a(X_a)] 
~,
$$
where the first term is the negative of the entropy and the second term is known as the average energy because $-\log f_a(X_a) = E(X_a)$. The free energy has a minimum value of $-\log(Z)$ and by minimizing this free energy we can also minimizing the KL divergence. 

We first consider the form of the free energy for a tree. For tree graphs, the distribution $q(X)$ can be written in the form:
$$
q(X) = \prod_{i} b_i(x_i)^{1 - d_i} \prod_{a} b_{a}(X_a) 
~,
$$
where the first product is over variables and the second is over the factors.
$b_i(x_i)$ is the marginal distribution over variable $x_i$, $b_a(X_a)$ is the joint marginal distribution over variables $X_a$ that connect to factor $f_a$, and $d_i$ is the degree of variable node $i$ (the number of nodes neighbouring node $i$). 
Plugging this into the expression for the entropy, we get: 
$$
H_{tree}(X) 
= -\sum_{i} (1 - d_i)  \sum_{x_i} b_i(x_i) \log b_i(x_i) - \sum_{a} \sum_{X_a} b_{a}(X_a) \log b_{a}(X_a) ~.
$$
Similarly, the average energy can be written as: 
$$
- \sum_{i} \mathop{\mathbb{E}}_q [\log f_i(X_i)] = - \sum_{a} b_a(X_a) \log f_a(X_a)
$$
Putting this together gives the free energy for tree graphs: 
$$
F_{tree} = - \sum_{i} (d_i-1) \sum_{x_i} b_i(x_i) \log b_i(x_i) + \sum_{a} \sum_{X_a} b_{a}(X_a) \log \frac{b_{a}(X_a)}{ f_{a}(X_a)} ~.
$$ 
For general factor graphs with loops, the $F \neq F_{tree}$.
The Bethe approximation is to use the free energy for a tree to approximate the free energy for arbitrary loopy graphs. The resulting approximate free energy is known as the Bethe free energy. 

Belief propagation can be derived via minimization of the Bethe free energy subject to two constraints.
The first is a marginalization constraint: $b_i(x_i) = \sum_{X_a \setminus i} b_{a}(X_a)$, and the second is a normalization constraint: $\sum_i b_i(x_i) = 1$.
With these constraints, we can form the Lagrangian and then set the derivates with respect to the parameters to zero: 
$$
L = F_{Bethe} + \sum_i \gamma_i \bigg\{ 1 - \sum_{x_i} b_i(x_i) \bigg\} + \sum_a \sum_{i \in N(a)} \sum_{x_i} \lambda_{ai}(x_i) \bigg\{ b_i(x_i) - \sum_{X_a \setminus i} b_{a}(X_a) \bigg\}
$$
$$
\frac{\partial L}{\partial b_{i}(x_i)} = 0 \;\;\;\;\;\;
\Rightarrow b_{i}(x_i) \propto \prod_{a \in N(i)} \exp \big(\lambda_{ai}(x_i) \big)
$$
$$
\frac{\partial L}{\partial b_{a}(X_a)} = 0 \;\;\;\;
\Rightarrow b_{a}(X_a) \propto f_{a}(X_a) \prod_{i \in N(a)}\exp \big( \lambda_{ai}(x_i) \big)
$$

We now choose the lagrange multiplier to be $\exp(\lambda_{ai}(x_i)) = m_{x_i \rightarrow f_a}(x_i) = \prod_{c \in N(i) \setminus a} m_{f_c \rightarrow x_i}(x_i)$. This is the familiar variable-to-factor message equation and substituting this into the above equations yields the belief propagation fixed point equations (the first of which the reader will recognize as the belief update rule).
$$
b_i(x_i) \propto \prod_{a \in N(i)} m_{x_i \rightarrow f_a} (x_i) \propto
\prod_{a \in N(i)} m_{f_a \rightarrow x_i}(x_i)
$$
$$
b_a(X_a) \propto f_a(X_a) \prod_{i \in N(a)} m_{x_i \rightarrow f_a}(x_i) = f_a(X_a) \prod_{i \in N(a)} \prod_{c \in N(i) \setminus a} m_{f_c \rightarrow x_i} (x_i)
$$

Using the marginalization condition , we can derive an equation for the messages in terms of other messages and produce the factor-to-variable message equation:
$$
\begin{aligned}
  m_{f_a \rightarrow x_i}(x_i) &= \sum_{X_a \setminus x_i} f_a(X_a) \prod_{j \in N(a) \setminus i} \; \prod_{b \in N(j) \setminus a} m_{f_b \rightarrow x_j}(x_j) \\
  &= \sum_{X_a \setminus x_i} f_a(X_a) \prod_{j \in N(a) \setminus i} m_{x_j \rightarrow f_a}(x_j)
\end{aligned}  
$$

This result tells us that the fixed-points of loopy belief propagation are local stationary points of the Bethe free energy and because the Bethe energy is bounded from below, BP always has a fixed point.

BP variants have been developed using more accurate or convex approximations of the free energy \cite{yedidia2000generalized}, however a detailed discussion of the theory behind BP is beyond the scope of this article and we refer the reader to \cite{wainwright2008graphical} for a in depth review.

\end{document}